\documentclass[manuscript, nonacm, table]{acmart}
\usepackage{xcolor}

\usepackage{graphicx}
\usepackage{amsmath}
\usepackage{booktabs}
\usepackage{CJKutf8}
\usepackage{pdfpages}

\usepackage{tabularx}
\usepackage{array}
\usepackage{soul}
\usepackage{tcolorbox}
\usepackage{multirow}
\usepackage[export]{adjustbox}

\usepackage[capitalize]{cleveref}
\crefname{section}{Sec.}{Secs.}
\Crefname{section}{Section}{Sections}
\Crefname{table}{Table}{Tables}
\crefname{table}{Tab.}{Tabs.}

\begin{document}

\title{The Surveillance AI Pipeline}
\author{Pratyusha Ria Kalluri$^*$}
\email{pkalluri@stanford.edu}
\orcid{0000-0001-7202-8027}
\affiliation{
  \institution{Computer Science Department, Stanford University}
  \streetaddress{353 Jane Stanford Way}
  \city{Palo Alto}
  \country{USA}
}

\author{William Agnew$^*$}
\email{wagnew3@cs.washington.edu}
\orcid{0000-0002-1362-554X}
\affiliation{% 
  \institution{Paul G. Allen School of Computer Science and Engineering, University of Washington}
  \streetaddress{185 E Stevens Way NE}
  \city{Seattle}
  \country{USA}
}

\author{Myra Cheng$^*$}
\email{myra@cs.stanford.edu}
\orcid{}
\affiliation{% 
  \institution{Computer Science Department, Stanford University}
  \streetaddress{353 Jane Stanford Way}
  \city{Palo Alto}
  \country{USA}
}

\author{Kentrell Owens$^*$}
\email{kentrell@cs.washington.edu}
\orcid{}
\affiliation{% 
  \institution{Paul G. Allen School of Computer Science and Engineering, University of Washington}
  \streetaddress{185 E Stevens Way NE}
  \city{Seattle}
  \country{USA}
}

\author{Luca Soldaini$^*$}
\email{lucas@allenai.org}
\orcid{0000-0001-6998-9863}
\affiliation{% 
  \institution{Allen Institute for Artificial Intelligence (AI2)}
  \streetaddress{2157 N Northlake Ave}
  \city{Seattle}
  \country{USA}
}

\author{Abeba Birhane$^*$}
\email{birhanea@tcd.ie}
\orcid{0000-0001-6319-7937}
\affiliation{% 
  \institution{Mozilla Foundation \& School of Computer Science and Statistics, Trinity College Dublin}
  \streetaddress{Dublin}
  \city{Dublin}
  \country{Ireland}
}

\begin{abstract}
\textbf{ABSTRACT}\\
A rapidly growing number of voices 
argue that AI research, and computer vision in particular, is powering mass surveillance. Yet the direct path from computer vision research to surveillance has remained obscured and difficult to assess. Here, we reveal the \textit{Surveillance AI pipeline} by analyzing three decades of computer vision research papers and downstream patents, more than 40,000 documents. 
We find the large majority of annotated computer vision papers and patents self-report their technology enables extracting data about humans. Moreover, the majority of these technologies specifically enable extracting data about human bodies and body parts. We present both quantitative and rich qualitative analysis illuminating these practices of human data extraction.
Studying the roots of this pipeline, we find that institutions that prolifically produce computer vision research, namely elite universities and ``big tech'' corporations, are subsequently cited in thousands of surveillance patents.  
Further, we find consistent evidence against the narrative that only these few rogue entities are contributing to surveillance. Rather, we expose the fieldwide norm that when an institution, nation, or subfield authors computer vision papers with downstream patents, the majority of these papers are used in surveillance patents.
In total, we find the number of papers with downstream surveillance patents increased more than five-fold between the 1990s and the 2010s, with computer vision research now having been used in more than 11,000 surveillance patents.
Finally, in addition to the high levels of surveillance we find documented in computer vision papers and patents, we unearth pervasive patterns of documents using language that obfuscates the extent of surveillance.
Our analysis 
reveals the pipeline by which computer vision research has powered the ongoing expansion of surveillance.

\def\thefootnote{*}\footnotetext{These authors contributed equally to the realization of this project.}\def\thefootnote{\arabic{footnote}}

\end{abstract}

\keywords{}
\maketitle

\section{Introduction}\label{sec:intro}
Over the past few decades, many groups, from grassroots communities to policymakers, have drawn attention to and organized against the rise of mass surveillance
~\cite{stoplapdspying, mijente, now2021ban,adalovelace,conger2019san}. 
Moreover, many have asserted that artificial intelligence (AI) research, and computer vision research in particular, is a primary source for designing, building, and powering modern mass surveillance
~\cite{monahan2018surveillance,lyon2010surveillance,scheuerman2021datasets,browne2015dark,agre1994surveillance,stark2019facial,zuboff2019age}. 
If these claims are true, the rapidly growing field of computer vision is contributing to the legacy of surveillance technologies that have exacerbated disparities, limited free expression, and created conditions that facilitated discrimination and abuse of power~\cite{Marx2015SurveillanceS, browne2015dark, monahan2018introduction, foucault1977discipline, deleuze1992postscript, adalovelace, allmer2011critical, richards2013dangers, stark2019facial}.  
Ascertaining the details of the pathways from computer vision to surveillance is urgent and crucial, as currently steep barriers prevent rigorously understanding, unmasking, and thus intervening on, these societally consequential technologies.
Here, we reveal the \textit{Surveillance AI pipeline}, illuminating the role of computer vision in surveillance.

\textbf{\textit{Computer vision}} refers to AI that focuses on measuring, mapping, recording, and monitoring the world from visual inputs such as image and video data. Computer vision has historical roots in military and carceral surveillance. As a technology that emerged in military contexts, it was historically developed to identify targets and gather intelligence in war, law enforcement, and immigration contexts~\cite{raji2021face,broussard2018artificial}. The field of computer vision now generally emphasizes training computers to interpret and understand the visual world. Yet, in-depth study of particular prominent computer vision tasks such as facial recognition has revealed that military history has heavily shaped core aspects and uses of these subfields. This further motivates interrogation into the extent to which the field of computer vision as a whole has been shaped in a way that powers mass surveillance.

\textit{\textbf{\textit{Surveillance}} at the most general level is defined as an entity gathering, extracting, or attending to data connectable to other persons, whether individuals or groups}~\citep{Marx2015SurveillanceS}. In the current computer age, surveillance is frequently ``extensive'':  entities, who are often minimally visible, use big datasets and aggregation to extend their reach, accessing previously unseen persons, locations, or information. 
Prominent examples are practices where entities in position of power observe, monitor, track, profile, sort, or police individuals and populations in private and public spaces through devices such as CCTV, digital traces on social network sites, or biometric monitoring of bodies~\cite{browne2015dark,monahan2018surveillance}. Through ubiquitously connected networks, data is aggressively gathered, shared, and aggregated. Behaviours, relationships and social and physical environments are datafied, modelled, and profiled. 
Many scholars emphasize that surveillance is inextricable from purposes such as influence, management, coercion, repression, discipline, and domination~\citep{allmer2011critical}. 
Crucially, a foundational understanding in surveillance studies is that technologies enabling the very possibility of monitoring suffice to foster conditions of fear and self-censorship, and this approach is a key means of social control~\cite{foucault1977discipline,deleuze1992postscript}.
We present a more extensive review of contextualizing literature in Appendix~\ref{sec:background}.

\begin{figure*}[!btp]
\caption{\textbf{Computer vision papers and downstream patents.} 
\\
This figure presents examples of computer vision papers and downstream patents, randomly drawn from our collected corpus of more than 40,000 such documents. For each paper and patent presented, an excerpt describing its goals and applications is shown, as well as an excerpted image. This provides a snapshot of our corpus}
\centering
\includegraphics[trim={0 5cm 0 -.5cm}, width=1\textwidth, page=1]{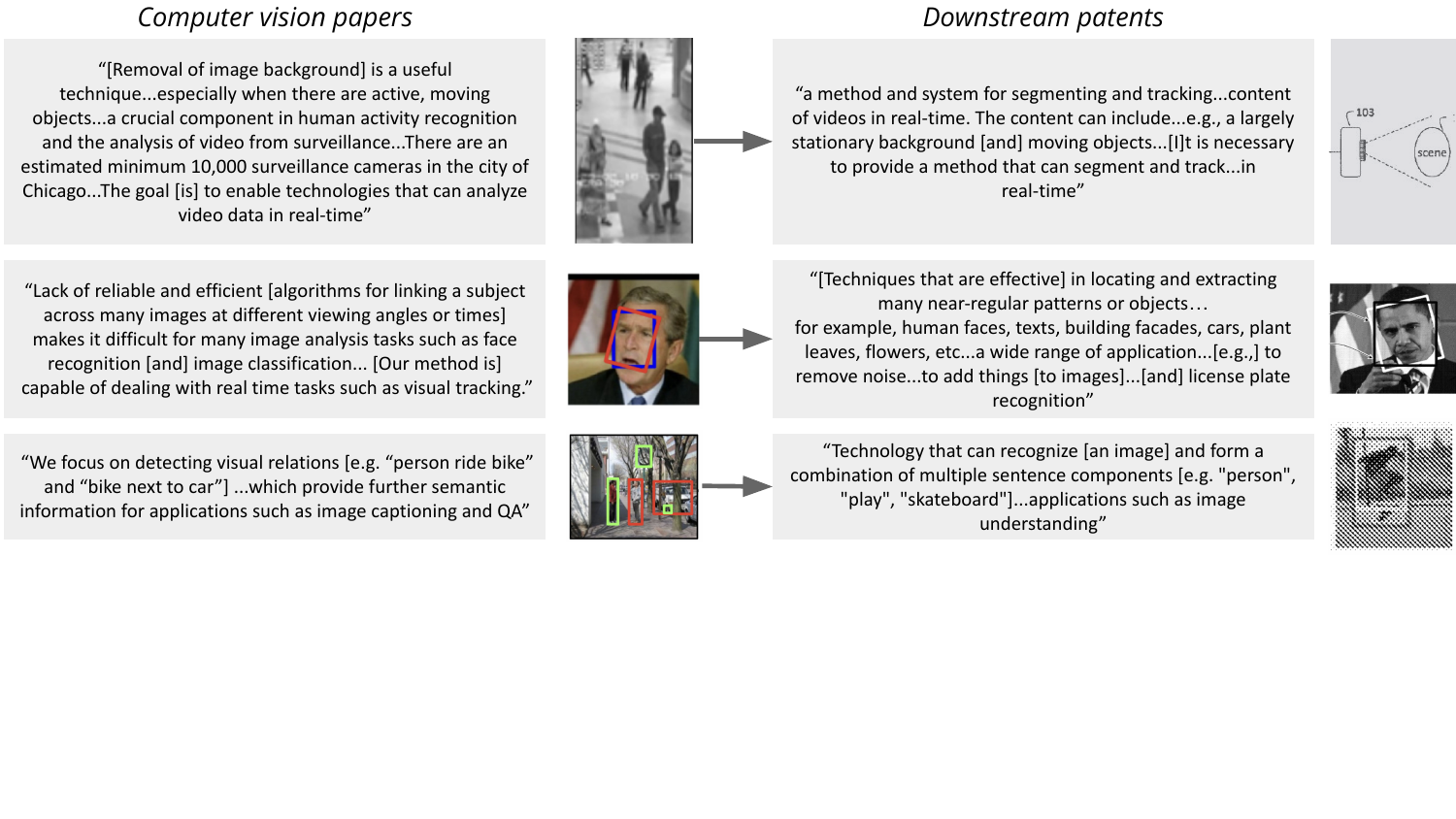}
\end{figure*}
\label{fig:randomexamples}

\textbf{\textit{The obfuscation of the Surveillance AI pipeline}} results from a confluence of forces. First, computer vision research is perceived by many as a neutral, purely intellectual endeavor, separate from downstream impacts and applications.
In fact, AI research at large rarely discusses connection to societal needs or potential negative consequences~\cite{birhane2022values}. 
Furthermore, surveillance often operates in the dark, and surveillance technology producers take extra measures to hide their existence~\cite{hill2020secretive,brewster}. 
It is difficult to gather direct evidence and details regarding the connections between research and surveillance applications: computer vision research papers and documentation, i.e., what research is being done, are written in ways that are not accessible to many outside the field; those who can parse this work are not accustomed or incentivized to elucidate the details of surveillance emerging; and research appears to trickle down in a complex multi-stage process. As a result, many aspects of the connection between computer vision research and surveillance remain shrouded in mystery.

\textbf{\textit{Our contributions}}. In this paper, an interdisciplinary team of researchers leveraged broad expertise including machine learning, AI, robotics, computer vision, privacy, science technology and society studies, history, AI ethics, and critical data studies to conduct an in-depth content analysis and large-scale computational analysis of three decades of computer vision papers and downstream patents. Notably, we study \textit{self-reported} uses and claims in the papers and patents. Thus, our findings are robust to claims of unintentional, unanticipated dual use by ``bad actors''; the extent and types of Surveillance AI we uncover in this paper are those that are intentionally indicated and fully anticipated by authors. Our key contributions are threefold:
\begin{enumerate}
\item\textbf{We reveal the pervasive extraction of human data}, quantifying the prevalence of extracting human data, identifying four  types of human data extraction, and presenting quantitative and rich qualitative analysis of the extraction taking place.
\item\textbf{We present a large-scale computational analysis of more than 40,000 computer vision papers and downstream patents to capture the roots and rise of Surveillance AI, across corporations, universities, nations, subfields, and years}. 
We reveal, for example, that it is not a few outlier institutions dedicated to surveillance. 
Rather it is a fieldwide norm that the majority of an institution's patented computer vision papers will be used in surveillance patents. 
Comparing the 1990s to the 2010s, these papers have become increasingly focused on surveillance, and the amount of computer vision used in surveillance has risen more than five-fold. 
\item\textbf{We shed light on widespread obfuscating language in papers and patents that contributes to perpetuating the paradigm of Surveillance AI.} This underscores the additional, hidden layers of interconnectedness between computer vision research and surveillance, above and beyond what is typically discussed.
\end{enumerate}

In making visible the pathways from computer vision research to surveillance applications, we aim for this mapping to serve as a tool for communities to strategically organize around and against surveillance; policy-makers to identify regulatory targets to curb surveillance; researchers to contend with the consequences of the field and (re)shape the research agenda; and the public to exercise the right to knowledge and power over the apps, gadgets, and devices that mediate and infiltrate their daily lives with 
surveillance.

\section{Corpus of computer vision papers and downstream patents}
To study the pathway from computer vision research to surveillance, we collect and analyze a corpus linking more than 19,000 computer vision research papers to more than 23,000 downstream patents. Research papers and patents have unique advantages making them revealing artifacts. First, they are primary sources written in researchers' and patenters' own words, with professional and institutional standards that they accurately describe their research and technologies and be able and willing to defend these documents' accuracy. 
The connections between research papers and citing patents serve as a rich data-trail of the path from research to applications~\cite{hammarfelt2021linking,ahmadpoor2017dual}.
We study papers published in the proceedings of 
the longest standing and highest impact computer vision conference, as it is highest impact by an extremely large margin; by standard h5-index these proceedings are among the top five highest impact publications \textit{in any discipline}, alongside Nature and Science. This research is widely seen to be an ``indicator of hot topics for the AI and machine learning community''~\cite{cvpr2021cvpr}. Acceptance and publication marks approval of research as work that exemplifies the core values of the computer vision community. As such, these papers both represent state-of-the-art in current computer vision and effectively reveal the values held in high regard within the community. 
We obtain all the proceedings published from 1990-2020 and, for each paper that has been cited in one or more patents, we obtain all citing patents. We refer to these as a \textit{patented paper} and its \textit{downstream patents}. 
Finally, we identify the which patents contain one or more surveillance indicator words, which we refer to as \textit{downstream surveillance patents}. We present additional methodological details in Section~\ref{sec:methodology}. In Figure~\ref{fig:randomexamples}, we present randomly sampled pairs consisting of a paper and a downstream patent, providing a snapshot of our corpus.

\section{The extraction of human data}
\label{sec:data}
There is extensive evidence of public distrust and fear concerning the capturing and monitoring of human data, including substantial concern about computer vision technologies operating on data ranging from online personal data traces to biometric and body data~\cite{adalovelace,nesterova2022questioning}.
To surface the potentially numerous and subtly expressed variants of human data extractions in computer vision, we conducted an in-depth qualitative content analysis of a subset of the corpus, analyzing one hundred computer vision papers and one hundred downstream patents (See Section~\ref{sec:methodology} for methodological details).
In the context of qualitative content analysis, this sample constitutes a large-scale manual analysis, which we complement with a large-scale computational analysis of the full corpus in subsequent sections.
This qualitative orientation is necessary when the key concepts that will emerge from a body of study are not known a priori, documents are complex or dense, expressing their key concepts with subtle language unique to the corpus, and a deep characterization is valuable~\cite{elo2008qualitative,vaismoradi2013content}. 
In this section, we present the resulting empirically grounded characterization, illuminating the types of data targeted in computer vision, along with to what extent parts or the whole of computer vision is explicitly dedicated to extracting human data.

{
\newcolumntype{R}[1]{>{\raggedleft\let\newline\\\arraybackslash\hspace{0pt}}m{#1}}
\setlength{\arrayrulewidth}{14pt} 
\setlength{\tabcolsep}{1em} 
\renewcommand{\arraystretch}{2}  
\begin{table}[]
\centering
\begin{tabular}{>{\centering}p{1.9cm} p{12cm}}
Human body parts 
& \cellcolor{gray!20}\textit{``The acquisition system may include a biometric sensor (e.g. an electronic fingerprint sensor, or an optical eye scanner, or a camera arranged to acquire a portrait image of an authorized person's face...'' (Patent 71)} \\
& \vspace{-20pt} {\small These technologies most frequently targeted faces, including
detection of eyes, eye movement, faces, “suspicious” facial expressions, and facial recognition. Some targeted other body parts, typically to enable human activity recognition.}
\\ 
\arrayrulecolor{white}\hline
Human bodies
& \cellcolor{gray!20}\textit{``...people monitoring in public areas, smart homes, urban traffic control, mobile application, and identity assessment for security and safety...'' (Paper 53)}
\\
& \vspace{-22pt} {\small These technologies most frequently targeted humans in the midst of everyday activities (e.g., walking, shopping, at group events) for purposes including body detection, tracking, and counting, as well as security monitoring and human activity recognition.}
\\
\arrayrulecolor{white}\hline
Human spaces
& \cellcolor{gray!20}\textit{``...a scene could be decomposed into a set of semantic objects...'' [accompanied by an example image taken inside an office] (Paper 40)}
\\
& \vspace{-22pt} {\small These technologies most frequently targeted living spaces, personal and communal, such as people's homes, offices, roads, town squares, or borders. The purpose was often identifying unspecified objects in these spaces; other specific purposes included modeling traffic and monitoring large border crossing areas.}
\\
\arrayrulecolor{white}\hline
{\multirow{2}{*}{
% X
\shortstack{Other\\socially salient\\human data}}}
& \cellcolor{gray!20}\textit{``Free-hand human sketches [e.g., of another person's item of clothing] are used as queries to perform instance-level retrieval of images'' (Paper 81)}
\\
& \vspace{-22pt} {\small These technologies targeted data containing traces of the mental, economic, cultural, social status, identities, preferences, or location details of humans, most frequently to narrow users' search results.}
\\
\arrayrulecolor{white}\hline
\end{tabular}
\caption{\textbf{Human data extraction in computer vision papers and patents. }
\\
In-depth content analysis identified four targets of human data extraction that were found in computer vision papers and downstream patents. 
These targets of extraction form a series of increasingly focused categories: % 
\textit{socially salient human data}, \textit{human spaces}, \textit{human bodies}, and \textit{human body parts}. This figure presents each category, with textual examples and qualitative description, and serves as the basis for our qualitative analysis.}
\label{tab:examples}
\end{table}
}

\begin{figure*}[]
\caption{\textbf{The prevalence of human data extraction in computer vision papers and patents.} \textbf{Left.} The large majority (90\%) of the annotated computer vision papers and patents enable extracting data about humans. The majority of the papers and patents (68\%) specifically enable extracting data about human bodies and body parts. Only 1\% of the papers and patents targeted only non-humans. (\textit{n=200, human bodies N=74 SD=6.7, human body parts N=62 SD=6.7, human spaces N=34 SD=5.6, unspecified N=22 SD=4.5, salient traces N=6 SD=2.3, non-human data N=2 SD=1.2}). \textbf{Right. }Figures in downstream patents make visible the prominence of targeting human bodies and spaces,
as is shown in this random sample of downstream patent images. We highlight those containing human bodies or body parts (\textit{in red}) and those containing human spaces (\textit{in orange}).}
\vspace{10pt}
\centering
\includegraphics[width=.9\textwidth, trim={0 6cm 0 0}, page=16]{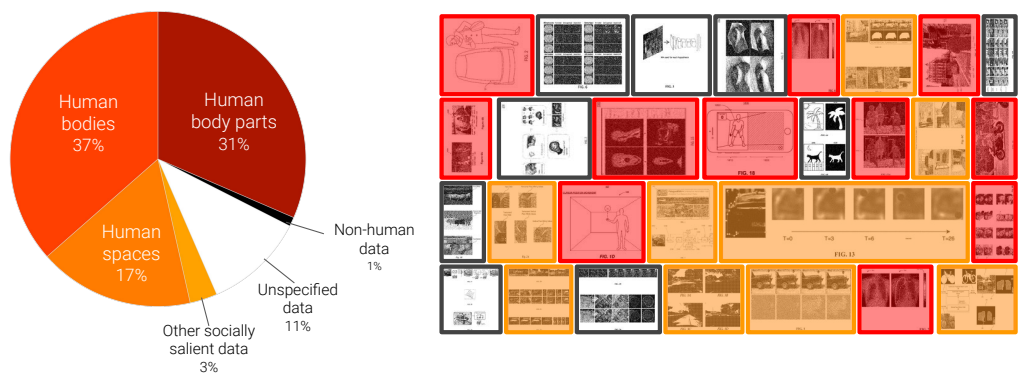}
\label{fig:data-type}
\end{figure*}

\subsection*{Quantitative analysis}

We present the types and extent of human data targeted in computer vision papers and downstream patents in Figure~\ref{fig:data-type}. In Figure~\ref{fig:data-type-split} we stratify this data to compare annotated papers versus patents. We find that 90\% of papers and patents extracted data relating to humans. Furthermore, the majority of papers and patents (68\%) explicitly extracted data about human bodies and body parts. In particular, at least a quarter of both papers and patents (35\% and 27\% respectively) claimed or demonstrated targeting human body part data as a strength of their technology, and at least an additional third of both papers and patents (36\% and 38\% respectively) claimed or demonstrated targeting human bodies. A smaller but still substantial portion of papers and patents (18\% and 16\%) extracted data about human spaces. 

Few papers and patents (1\% and 5\%) present their technology as useful for monitoring, tracking or predicting non-body-related socially salient human data. Strikingly, only 1\% of papers and patents were dedicated to extracting only non-human data, revealing that both computer vision research and its applications are overwhelmingly concerned with datafying humans and specifically human bodies. Finally, the remaining portion (11\%) of papers and patents claimed to capture and analyze ``images'', ``text'',  ``objects'', or similarly generic terms, leaving unstated whether they anticipated these categories including humans or human data.

\subsection*{Qualitative analysis}
Our guiding aim was to cast light on the nature of the dense bodies of computer vision research and applications and elucidate connections to surveillance, especially as these papers and patents can each be dozens of pages, difficult to obtain and link, and written in a manner that assumes the reader has substantial expertise in computer vision, disciplinary jargon, academic research, and patent applications. Here, on the basis of our in-depth content analysis, we present the four targets of human data extraction that were identified in computer vision papers and patents, alongside examples and qualitative description (Table~\ref{tab:examples}).  
Further, we find that our analysis responds to narratives that only a minute portion of computer vision and data extraction are harmful, and many kinds of computer vision and data extraction are benign or harmless. Rather, we find that computer vision prioritizes forms of data extraction that are widely viewed as the most intrusive, and, drawing from the large body of surveillance studies scholarship, we see that other forms of prominent data extraction are not less intrusive -- they merely intrude in distinct ways.

The four targets of human data extraction that emerged form a series of increasingly focused categories: % 
\textit{socially salient human data}, \textit{human spaces}, \textit{human bodies}, and \textit{human body parts}. Papers and patents broadly assumed tasks targeting human body part data as valuable, particularly targeting facial analysis, and sometimes enabling activity classification. This validates the substantial concerns that have been put forth regarding biometric and related body part data. Biometrics such as faces, fingerprints, and gait, which constitute uniquely personal data that is often inseparable from our identities, has proliferated as a form of surveillance in recent years, from fingerprint detection to activity recognition technologies that emphasize tracking body parts.
Their pervasiveness has been shown to significantly infringe on people's privacy and threaten human rights~\cite{adalovelace}. 

Further, the papers and patents targeting human bodies most frequently targeted humans in the midst of everyday activities (e.g., walking, shopping, at group events), and named purposes included body detection, tracking, and counting, as well as security monitoring and human activity recognition. The dominance of analysis of human bodies in everyday settings aligns with the view of new surveillance by~\citet{browne2015dark} who % 
characterizes the new practices of surveillance as often \textit{undetected} -- for example cameras hidden in everyday benign objects -- or even \textit{invisible}. In these forms, data is frequently collected without consent of the target, and then shared, permanently stored and aggregated. \citet{browne2015dark} characterizes surveillance as focused on monitoring and cataloguing that which was previously left unobserved, with the human body as a primary site of surveillance.

Beyond human bodies, analysis of human spaces was widespread in papers related to scene analysis, understanding, or recognition, which are often presented as a core contribution of the field in papers and patents alike. These technologies targeted personal and communal living spaces, including people's homes, offices, roads, town squares, or borders. Purposes were often generic, identifying unspecified objects; more specific purposes included modeling traffic and monitoring  borders. This is one way of making previously unobserved phenomena, events, interactions and places amenable to observation, which~\cite{cohen2017surveillance} document as a fundamental mechanism of surveillance. The rendition of homes, streets, neighbourhoods, villages and towns to surveillance technology marks these spaces as no longer scenes where residents, live, meet and talk but another object of target for data collection, tracking, categorizing, and predicting~\cite{zuboff2019age}. The consequence of the gradual rendering of more and more of these spaces is extremely subtle yet has profound implications for the future of humanity. It accumulates to what Zuboff calls the condition of ``no exit'', where there are fewer and fewer spaces left to ``disconnect'', seek respite and be left to just be~\cite{zuboff2019age}.
Similarly, capturing socially salient human data contributes to the gradual cataloguing, documenting, mapping, and monitoring of human affairs in its rich complexities~\cite{monahan2018surveillance,zuboff2019age}. 

Finally, through close inspection, we find that the targeting of generic or unspecified data does not imply that the technology described in the paper or patent cannot be used on human-related data or even that human data was not a desired use case by the authors. In Section~\ref{sec:obfuscating_lang}, we present evidence that, to the contrary, we find dense patent language can hide the human data analysis in the upstream papers and, conversely, papers that do not speak to the potential for use with human data often lead to patents that explicitly monitor human data, contributing to an additional obfuscated layer of human data extraction. These findings provide analysis and concrete examples challenging the casting of human data extraction as frequently benign, as we see that the forms of data extraction present are each highly intrusive and distinct in the manner of intrusion.

\section{The rise of Surveillance AI}
\label{sec:rise}
\begin{figure*}
\caption{\textbf{The rise of computer vision for downstream surveillance.}\\
\textbf{Left.} Across three decades of patented computer vision papers (\textit{n=11,917}), there has been a steady increase in the proportion used in surveillance patents. In the 1990s, only one half of patented computer vision papers were used in surveillance patents (\textit{percent=50\%, SD=2\%, n=664}), yet in the 2010s, a large majority of patented computer vision papers were used in surveillance patents (\textit{percent=79\%, SD=1\%, n=2,327}). \textbf{Center.} Comparing the 1990s to the 2010s, the number of computer vision papers used in only non-surveillant patents has remained relatively stable, while the number of computer vision papers used in surveillance patents has risen more than five-fold. Whiskers represent standard deviation. \textbf{Right.} To assess linguistic evolution of computer vision papers across decades, we measure differences in word frequency between 1990s paper titles versus 2010s paper titles. We report highly polarized words (with z-scores computed using weighted log-odds ratios). There is a clear qualitative shift from more generic paper focus in the 1990s (\textit{teal bars}) to an increased focus on analysis of semantic categories and humans (e.g. “semantic”, “action”, “person”) in the 2010s (\textit{pink bars}). All shown word associations are statistically significant ($p< .01$).
}
% \centering
\includegraphics[page=5, height=6cm, trim={2cm 6.5cm 8cm 0cm, clip, center}]{latex/images/figs.pdf}
\label{fig:years}
\end{figure*}

\let\includegraphicsold\includegraphics

\begin{figure*}
\caption{\textbf{The fieldwide dominance of downstream surveillance.} \\
\textbf{Top.} For the top institutions and countries authoring the most computer vision research, we show the large number of these papers subsequently used in surveillance patents. For every one of these prolifically publishing institutions and countries, we see the majority of its patented papers are used in surveillance patents. \textit{(Teal bars are larger than grey bars.)} \textbf{Bottom.} This aligns with a computer vision fieldwide norm. For each institution, country, and subfield that have published at least 10  papers with downstream patents, we show the percent of these papers that are used in surveillance patents \textit{(vertical grey bars)}  (\textit{n=13,804}, \textit{n=18,272}, \textit{n=19,413}). We find a pervasive norm: \textit{when an institution, nation, or subfield authors papers with downstream patents, the majority are used in surveillance patents}. (\textit{Vertical grey bars are consistently above the orange 50\% threshold}.) Whiskers represent standard deviation. The name of each individual entity and its corresponding percentage (\textit{i.e. the label of each individual vertical grey bar)} is included in the Appendix for further inspection. }
\vspace{10pt}%
\centering
{\begin{minipage}{\textwidth}
\hspace{.02\textwidth}%
\includegraphics[height=110px]{latex/images/top_insts.pdf}
\includegraphics[height=110px]{latex/images/top_countries.pdf}
\vspace{.6cm}
\end{minipage}}
% \rule{10cm}{.4pt}
\centering
\begin{minipage}{\textwidth}
\includegraphics[width=105px, trim={0 7cm 0 0}, clip, center]{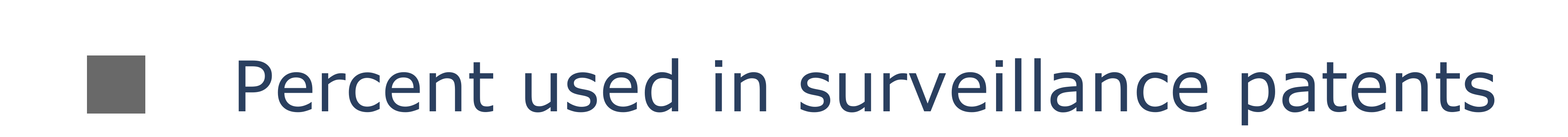}
\end{minipage}
\centering
\begin{minipage}{.8\textwidth}{
\includegraphics[height=40px]{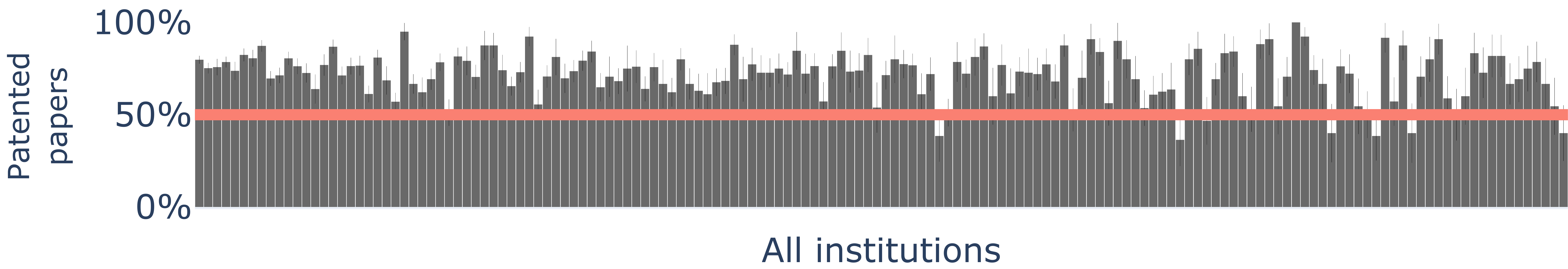}%
\includegraphics[height=40px, trim={5cm 0 0 0}, clip]{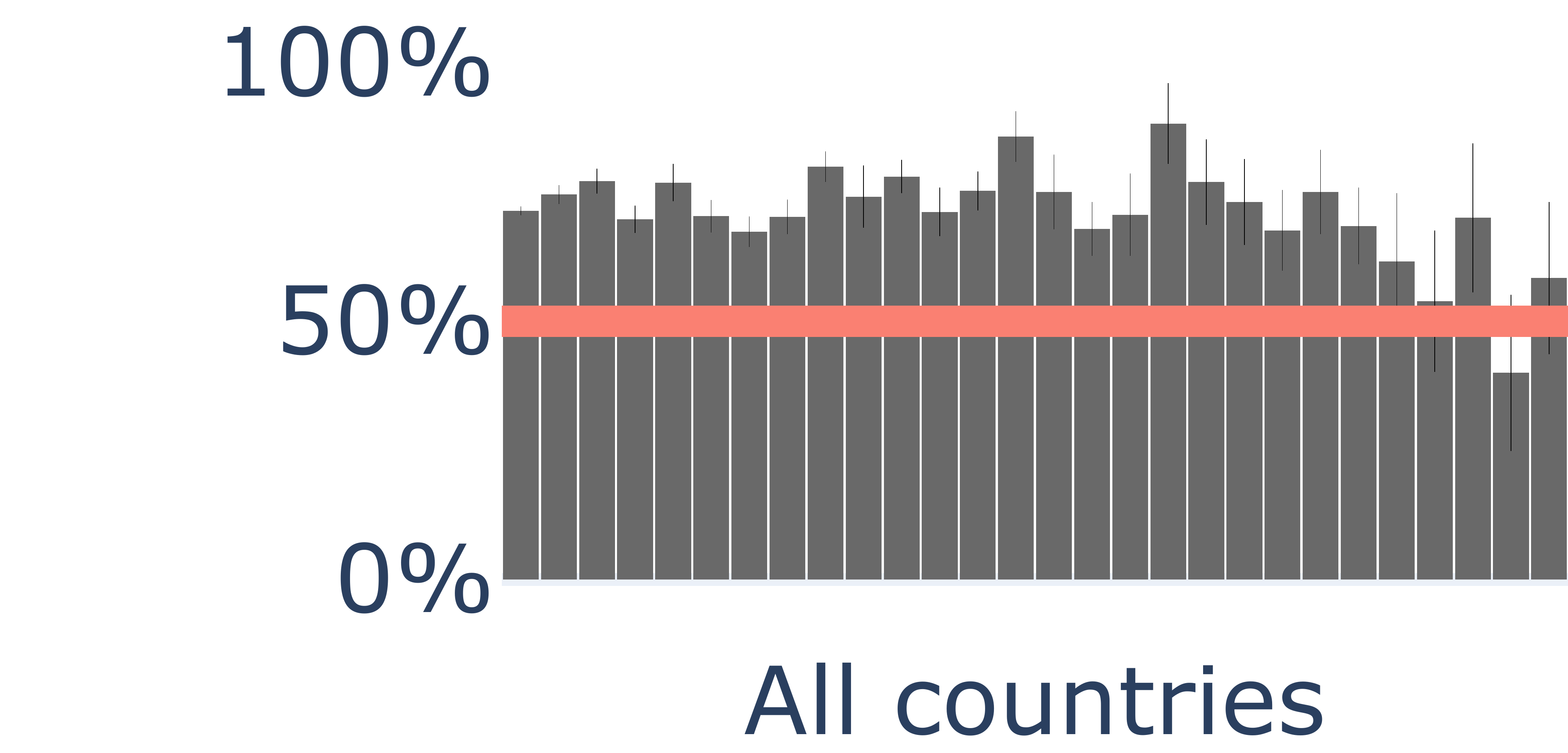}}
\vspace{.2cm}
\end{minipage}
\centering
\begin{minipage}{.75\textwidth}
% \hspace{57px}
\includegraphics[height=40px,]{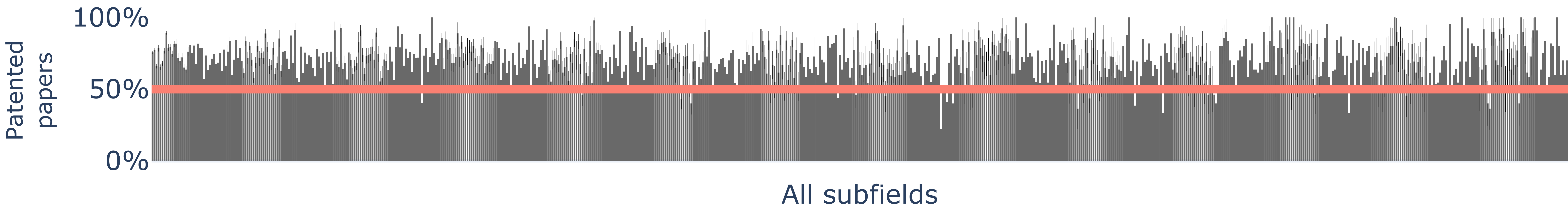}
\end{minipage}
\label{fig:institutions}
\end{figure*}

We present the evolution of computer vision papers and downstream patents in Figure~\ref{fig:years}. We find a substantial increase in papers used in surveillance patents. Comparing decades, we find that the 1990s produced relatively fewer computer vision papers with downstream patents, and only half of these  were used in surveillance patents (\textit{percent=50\%, SD=2\%, n=664}). Two decades later, the 2010s show a tripling of computer vision papers with downstream patents, and 79\% were used in surveillance patents (\textit{percent=79\%, SD=1\%, n=2,327}). The twin forces of the increase in computer vision papers with downstream patents and the increase in the proportion of these used in surveillance patents combined to large effect: 
the 1990s to the 2010s constituted a more than five-fold increase in the number of computer vision papers used in surveillance patents (Figure~\ref{fig:years}, \textit{center}). 

It is also possible to gain insight into the evolution of computer vision by inductively surfacing patterns of linguistic evolution. To study the linguistic evolution that has occurred over the past several decades, we compare the log-odds ratios of word frequencies in 1990s paper titles versus 2010s paper titles. We use the informative Dirichlet prior to obtain measures of statistical significance and control for variance in words’ frequencies \citep{monroe2008fightin}. In Figure~\ref{fig:years}, we show highly polarized word associations in both directions with computed z-scores. Additional methodological details are presented in Appendix~\ref{sec:methodology}. We see a clear, qualitative shift from more generic application-ambiguous language in the 1990s (e.g. ``shape'', ``edge'', ``surfaces''; \textit{teal bars}) to an increased focus in the 2010s on analysis of semantic categories and humans (e.g. ``semantic'', “action”,“person”; \textit{pink bars}). Due to this being an inductive analysis that surfaces the dominant patterns of linguistic evolution, this finding suggests that not only is there a major change toward rising surveillance but it is one of the most salient changes that have occurred in the field over the past several decades. Taking our results together, we see that the language and patenting practices in computer vision have evolved in ways that increasingly focus on analyzing humans and enabling surveillance.

\section{The normalization of Surveillance AI}
\label{sec:whois}
Surveillance technology does not emerge in a vacuum.
Over time, researchers from many nations, institutions, and subfields have conducted this research and developed applications, whether cognizant of and attentive to its downstream consequences or not.
This work has been actively funded and commercialized by external parties, and it continues to evolve. 
In Figure~\ref{fig:institutions} (\textit{top)}, we show the institutions and nations authoring the most computer vision research, and we present the large number of these papers subsequently used in surveillance patents. For every one of these prolifically publishing institutions and nations, we see the majority of its patented papers are used in surveillance patents (\textit{Fugure~\ref{fig:institutions} top, teal bars are greater than grey bars}).
These top institutions are often prestigious institutions, including ``big tech'' corporations and elite universities, many of which are also the top producers of computer science papers generally~\citep{Berger_CSrankings_2017, ahmed2020dedemocratization}, reflecting tight ties between those driving research and those driving surveillance. 
The prevalence of both elite universities and major tech corporations in Figure~\ref{fig:institutions}, along with an across-the-board practice of being used in surveillance patents, reflects the research ties between universities and corporations that have shaped the field of computer science from its nascence~\citep{edwards1996closed}.  
The institutions authoring the most papers with downstream surveillance patents align with well-established historical legacies of the military-industrial-academic complex \citep{leslie1993cold, ziegler1995spying, abbate2000inventing, leffler2010cambridge, ahmed2015cia, crawford2021atlas}.

To understand the influence of nations and their research output on surveillance patents, we additionally present in Figure~\ref{fig:institutions} the distribution of ties to surveillance across authoring countries.
The authoring countries are obtained from the location of paper authors' institutional affiliations.
The top two nations producing papers with downstream surveillance patents are the US and China by a large margin, with the US producing more of these papers than the next several nations combined.
Our findings correspond to previous reports about AI-driven surveillance across countries, which state that on a global scale, China and the United States are the major drivers in supplying advanced surveillance technologies, while the major users include both liberal democracies and other countries with less democratic governments~\citep{feldstein2019global}. We connect these statistics to the narrative of ongoing tensions between the United States and China to establish themselves as the global superpower in AI~\citep{state2017next}. From this perspective, the institutional race to develop surveillance technologies is cast as a mission to defend against an enemy and align with state agendas~\citep{chun2006control,crawford2021atlas}.

These findings provide the basis for a salient question: are only a few rogue entities contributing to surveillance or are ties from research to surveillance a fieldwide norm? 
We find substantial evidence against the narrative of only a few rogue entities contributing to surveillance. 
Rather, 
we identify a pervasive norm: \textit{when an institution or nation authors computer vision papers with downstream patents, the majority are used in surveillance patents.} (\textit{Figure~\ref{fig:institutions} bottom, institutions' and nations' vertical grey bars are consistently above the orange 50\% threshold.}) 
This norm describes the behavior of
74\% of institutions and 83\% of nations, evidencing the wide-spanning normalization of computer vision used in surveillance.
Similarly, we find substantial evidence against the narrative that there are merely a few implicated subfields of computer vision within a broader non-surveillance-oriented field. Rather, we find the continuation of the norm: \textit{when a subfield produces computer vision papers with downstream patents, the majority are used in surveillance patents}. (\textit{Figure~\ref{fig:institutions} bottom, the majority of subfields' vertical grey bars are above the orange 50\% threshold.}) Inspecting the computer vision subfields that author papers with downstream patents, it may be expected that the stated norm describes frequently implicated subfields such as facial recognition, but in fact we find that the norm describes the majority (69\%) of the subfields. Our findings indicate that, across institutions, nations, and subfields, the practice of producing computer vision that enables surveillance is a pervasive fieldwide norm.

\section{The obfuscating language of Surveillance AI}
\label{sec:obfuscating_lang}
Finally, in addition to our analysis of self-reported ties to surveillance, we illuminate striking trends of obfuscating language that minimized or sidestepped mentions of potential surveillance and discussion of its harms. We highlight and qualitatively describe two salient themes that emerged: 

\begin{center}{\textbf{1. Papers and patents cast humans as merely another entity under the umbrella term ``objects''.}}

\vspace{10pt}
\textit{``We will simply use the term objects to denote both interactional objects and
human body parts'' (Paper 84)}

\textit{``Using these methods, objects such as people and vehicles may be identified and quantified based on image data.'' (Patent 85)}

\textit{``Since the surveillance system detects and can be interested on vehicles,
animals in addition to people, \\hereinafter we more
generally refer to them with the term moving object.'' (Paper 53)}

\end{center}
\vspace{5pt}

\noindent Establishing the conceptualization of human as merely a kind of object explicitly, as many papers and patents do, enables the rest of those documents and, crucially, \textit{all other papers and patents} to merely discuss problems related to \textit{objects} or \textit{scenes}, as they can rely on the understanding of human as object that has been established by peers. Because humans are considered objects and scenes often contain people, such abstractions indicate that the field understands any paper or patent that discusses objects and scenes -- which constitutes the majority of the field -- as potentially enabling surveillance of humans.
Many papers conflate humans with objects, making no note of how performing tasks like detection or segmentation on people has extremely specific, and socially consequential impacts. For instance, a paper about panoptic segmentation, in giving context about the body of literature that it draws from, makes no distinction between non-human detection and face detection: ``Early work on face detection...helped popularize bounding-box object detection. Later, pedestrian detection datasets helped drive progress in the field'' (Paper 96). The lede of a paper about parsing object interactions does the same: ``a major task of fine-grained interaction action analysis
is to detect the interacting objects or human body parts for
each video frame (in the rest of the paper, we will simply
use the term objects to denote both interactional objects and
human body parts)'' (Paper 84).
Considering humans as objects implies that any knowledge produced related to object-focused tasks can be directly applied to human data. This assumption neatly abstracts away the ways that such methods can be applied to surveillance. This phenomenon also ties to literature about traditional science’s sharp divide between subject and object, which positions scientists as % 
the studiers of ``objects'' out there. % 
This ``splitting of subject and object'' facilitates ``denial of responsibility and critical inquiry''~\citep{haraway2020situated}. This contextualizes the field's homogenization of all possible data, including human data, into objects to be studied, often without consent and without consideration of their sources or consequences.

\begin{center}
\textbf{2. \textit{What is not said}: Even when the text of papers and patents makes no mention of human data, \\figures or datasets may contain many, sometimes exclusively, images of humans.}
\end{center}
The pattern of papers and patents claiming to target ``objects'', while briefly defining these terms as subsuming humans, sets a clear precedent that we find has already played out: we find that other documents lean on these norms, claim to target ``objects'', in actuality target humans, and thus leave \textit{no textual trace} of the human data extraction they are engaged in. For example, one paper describes itself as improving object classification and makes no mention of humans; yet close inspection of the paper's first figure reveals (in 3-point font) that it classifies so-called objects into classes including ``person'', ``people'', and ``person sitting'' (Paper 5). A second paper describes itself as identifying salient regions of images and does not mention humans in its text or figures; yet inspection of the paper's datasets reveals they demonstrate their technology by detecting regions of interest such as humans walking on a sidewalk (Paper 1).  
Figure~\ref{fig:data-type} illuminates this pattern via a random sample of images drawn from downstream patents, many of which we find contain human bodies and spaces despite many lacking explicit mention of these entities in the text.
This further entrenches a field norm that, on one hand, humans and objects of all kinds may be targeted in parallel, casting the vastly different implications as inconsequential, and, on the other hand, that humans can be central targets of technologies without needing to leave a textual trace, let alone discuss, surveillance. This norm obscures the extent of Surveillance AI from both outsiders attempting to understand the field and insiders not cognizant of these practices. In this way, the modeling and categorization of humans has become so pervasive it can only be understood as a task that has become widely acceptable across the field of computer vision, rendering it a potential application of virtually all computer vision papers and patents.

\section{Discussion: A Paradigm of Surveillance}\label{sec:paradigm go}

The studies presented in this paper ultimately reveal that the field of computer vision is not merely a neutral pursuit of knowledge; it is a foundational layer for % 
\textit{a paradigm of surveillance.}
Our findings include these striking points: 
90\% of papers and patents emphasize it as a strength that their technologies can target human data. 
Not only is human data broadly targeted, but the \textit{majority} (68\%) of papers and patents explicitly focus on surveillance of human body parts (e.g., faces) and human bodies. Between the 1990s and 2010s, we have seen the rise of Surveillance AI, and it has become an overwhelming norm that computer vision papers analyze humans, and those papers used in patents are most likely used in surveillance patents.
Moreover, even when a paper does not \textit{explicitly} state surveillance as an application, it provides the methods to do so and is grounded in a historical context that makes it possible to target human surveillance while minimizing the acknowledgement of these intentions.
In other words, the default stance of the field ties progress closely with surveillance. This is evident in the types of research questions that are valued and prioritized by the field, as well as the way that the papers are written -- particularly the use of obfuscating language -- for example the use of ``object'' in the analysis of humans, sometimes only exposing the anticipated human data in images and figures.
 
The uncovered features of computer vision tie into a broader literature about the veneer of neutrality in science. Scientific findings are frequently falsely presented as facts that emerge from an objective ``view from nowhere'', in a historical, cultural, and contextual vacuum. % 
Such views of % 
science as “value-free” and “neutral" have been debunked by a variety of scholarships, from philosophy of science, STS and feminist and decolonial studies. A purported view from nowhere is always a view from somewhere and usually a view from those with the greatest power~\citep{proctor1991value,harding2013rethinking,haraway2020situated,longino2020science}. % 
Social and cultural histories and norms, funding priorities, academic trends, researcher objectives, and research incentives, for example, all inevitably constrain and shape the direction and production of scientific knowledge
~\citep{abbate2012recoding,ensmenger2012computer,ensmenger2015beards, birhane2022values}.
An assemblage of social forces have shaped computer vision, resulting in a field that now fuels the mass production of Surveillance AI.

Peering past the veneer of scientific neutrality, we find that the ongoing expansion of the field of computer vision is centrally tied to the expansion of Surveillance AI. 
At its core, % 
surveillance is the perpetual practice of % 
rendering visible what was previously shielded and unseen~\citep{browne2015dark}. This is precisely the goal % 
of the discipline of computer vision. % 
The continued progress of the field amounts to % 
increasing the capabilities for recording, monitoring, tracking, and profiling of humans as well as the wider social and physical environment. 
These tasks, which may seem benign to those swimming in the waters of computer vision, in fact exemplify the ways that progress in the field of computer vision is inextricable from increasing surveillance capabilities. % 

Ultimately, whether a work in the field of computer vision predicates surveillance applications or not, it can and frequently will be used for these purposes. Given the ways that research throughout the field can be implicated and engaged in surveillance, even when the precise details are missing or obfuscated, our findings may constitute a lower bound on the extent of computer-vision based surveillance: there are likely many more works that have quietly contributed to Surveillance AI. Viewing computer vision in this light, it becomes clear that shifting away from the violence of surveillance requires, not a small shift in applications, but rather a reckoning and challenging of the foundations of the discipline.

\section{Methodology}
\label{sec:methodology}

\subsubsection*{\textbf{Data}}
In addition to the unique advantages of analyzing papers we describe in the main text, additional advantages include that they must report their authors, primary affiliated institutions, and years of publication, enabling reliable analysis of how these factors influence the pathway to applications; they are available online; and they have a consistent overall structure facilitating consistency of annotation and reliable comparisons. These papers and their collected downstream patents served as the basis for the in-depth content analysis and large-scale automated analysis presented in this paper. We study the longest standing and highest impact (by an extremely large margin) computer vision conference, which is the Conference on Computer Vision and Pattern Recognition (CVPR). Throughout our studies, we analyze the corpus of CVPR papers from 1990-2020. In 1990, 1995, and 2002, CVPR did not occur, so there are no papers from these years. 
In constructing our corpus, we leverage and link the papers in the Microsoft Academic Graph~\cite{sinha2015overview}, the paper-patent citation linkages inferred by ~\citet{marx2022reliance}, and the patents in the Google Patents database. 
Manual verification found the paper-patent citation linkages to have over 99\% precision and 78\% recall.
All papers at CVPR are published in English. For patents that were published in other languages, the Google patents English translations were used.

\subsubsection*{\textbf{Content analysis}}
Following best practices in content analysis, we conducted an in-depth analysis of a purposive sample of papers and patents distinctively informative of the development of computer vision research and applications. 
For each year from 2010 to 2020, we randomly sampled ten paper-patent pairs that consisted of a CVPR research paper published in this year and a downstream patent. This formed a total of 100 papers and 100 downstream patents. In the context of content analysis, this constitutes a large-scale annotation.

We conducted the content analysis using close reading of documents and a rigorous qualitative methodology. An interdisciplinary six-person team analyzed the documents using an integrated inductive-deductive methodology. In the inductive component, each document was read line by line including figures, inductively coding key emergent features of the technology's treatment of human data and iteratively accumulating a list of these key features and their relationships.
We complemented this with an additional deductive component in order to ensure that we actively looked for and captured instances of papers and patents with key features that inhibited usage for surveillance, even if rare. In this deductive component, we coded for two such features. The inductive and deductive codes are discussed in this section, Section~\ref{sec:data}, and Appendix ~\ref{sec:transfer} and~\ref{sec:app-use}.
 
 Our annotation team had several strengths: our team included both published experts in computer vision and field outsiders, allowing for expert insights and translation, as well as fresh perspectives that could illuminate computer vision disciplinary biases. 
 We utilized the constant comparative method.
Throughout the coding process the team held frequent, extensive discussions to develop the precise meanings of codes and their relationships, and to revise and refine the code list. 
At the end of all coding, the team unanimously agreed upon the key emergent dimensions and features, along with the relationships amongst these dimensions and features, which we summarize in Figure~\ref{fig:topology}, and discuss in detail in Section~\ref{sec:data} and Appendix~\ref{sec:transfer} and~\ref{sec:app-use}.
Additionally, as we coded papers and downstream patents we encountered and discussed salient examples of \textit{obfuscating language} being used to describe or avoid describing surveillance, and we present these findings in Section~\ref{sec:obfuscating_lang}.

On the basis of our in-depth, interdisciplinary content analysis, we present the Surveillance AI topology in Figure~\ref{fig:topology}, bringing to the fore the dimensions, features, and dynamics of computer vision's treatment of human data and connecting these to concepts in surveillance studies that elucidate the complexity and consequences of these particular findings. Our analysis identified three key dimensions capturing these technologies' treatment of human data: \textit{(1) Data type} — What type of data does the technology extract, attend to, capture, monitor, track, profile, compute, or sort, and to what extent is it human and personal? \textit{(2) Data transferal} — To what extent does the data remain under the control of the datafied person or become transferred to others? \textit{(3) Use of data} — For what purpose is the data used? These three dimensions are discussed in detail, with examples and analysis, in Section~\ref{sec:data} and Appendix~\ref{sec:transfer} and~\ref{sec:app-use}.

We discuss the primary dimension of the topology in detail in Section~\ref{sec:data}. In this primary dimension, the inductively identified types of human data extracted form a series of nested, increasingly focused categories: \textit{socially salient human data}, \textit{human spaces}, \textit{human bodies}, and \textit{human body parts}. A fifth inductively identified target of data extraction was \textit{General / unspecified data}, which tended to target generic tasks such as ``identifying objects'', did not specify targeting human data, but also did not commit to targeting only non-human data. In addition to these data types, which were inductively found only through close reading of the papers and patents, the annotation team deductively included \textit{non-human data} in the annotation scheme from the start. This was to ensure that we captured mentions of any non-surveillance technologies in papers and patents, even if rare. In order to enable a quantitative analysis of this primary dimension, for each paper and patent, we identified the innermost (most focused) type of human data extracted. Half of the documents were annotated by more than one annotator, which was particularly valuable for becoming accustomed to types of cases in which a single sentence or figure influenced the appropriate code; the existence of such cases is discussed in Section~\ref{sec:obfuscating_lang}; in these cases of multiple annotators, each document's final code was determined through discussion until consensus. We then quantized all documents' annotations, presenting the relative frequencies of the data types in Figures~\ref{fig:data-type} and~\ref{fig:data-type-split}. 
The second and third dimensions of the topology are less consistently discussed in papers and patents. Nonetheless, key areas of surveillance studies scholarship are dedicated to how these dimensions (data transfer and data use) are important to understanding the roles, dynamics, and consequences of surveillance. Given the importance of these dimensions, in Appendix~\ref{sec:transfer} and~\ref{sec:app-use} we include a full discussion of these dimensions, the inductive and deductive codes, demonstrative examples and findings, and connections to nuanced dynamics of surveillance that have been discussed in surveillance studies literature.

\subsubsection*{\textbf{Automated analysis}}
 To study the breadth and variation of surveillance across years, institutions, nations, and subfields, we 
 conducted a large-scale computational analysis of more than 40,000 papers and patents. Specifically, during the in-depth manual content analysis the team of annotators identified a list of surveillance indicator words that indicated surveillance (in particular, words that indicated the targeting of human body parts, human bodies, human spaces, or socially salient human data; Section~\ref{sec:data} provides detailed discussion of each of these types of targeting and discussion of how they enable surveillance). To validate each surveillance indicator word, we scanned the corpus for all patents containing this word, randomly sampled ten of these patents, and conducted manual inspection. We removed from the list all words that manual inspection identified as not reliable indicators (typically because they had frequent additional word senses; e.g. a ``store'' could be a human space but was frequently a technical term related to data or memory storage, so was removed from the list). The resulting list of surveillance indicator words was approved by consensus, and we list these words in Appendix~\ref{ref:keywords}.

For each paper, we scanned its downstream patents to identify patents containing one or more of these surveillance indicator words, which we refer to as downstream surveillance patents.
We present the distribution of surveillance patents across institutions, nations, subfields, and years, along with contextualizing discussion, in Sections~\ref{sec:rise}~and~\ref{sec:whois}.
We present additional methodological details in Appendix~\ref{sec:method-extra}. 
\subsubsection*{\textbf{Analysis of the evolution across years}}
To conduct an analysis across years (displayed in Figure~\ref{fig:years}), we filter the corpus years. In emerging and developing fields, the estimated time from a paper being published to a downstream patent being published is three to four years; this is the time from the paper being published to the downstream patent being filed as well as the time of the patenting process~\cite{finardi2011time}. This appears to be in line with our corpus, as the number of computer vision papers with downstream patents stabilized in the early 2000s and from the early 2000s onward remained above 200 every year, until 2018 (exactly four years before our analysis began), at which point it suddenly dropped by nearly a half. Accordingly, for the analysis across years, we removed papers from the years 2018 and 2019 since these were less than four years before our analysis began so many papers had not yet had the chance to have the majority of their patenting process play out, leading to less reliable analysis. This filter had the added benefit that, in our analyses comparing the 1990s to the 2010s, both decades consisted of 8 years, putting these decades on a fair playing field for totaling when comparing the number of downstream patents of various types.

To study the linguistic evolution that has occurred, we compute the log-odds ratio with Dirichlet prior of words appearing in 1990s paper titles versus 2010s paper titles \citep{monroe2008fightin}. We remove stop-words (those in the NLTK stopwords list, as well as ``using" and ``via", because ``using" and ``via" are common stopwords in computer vision titles). We then present ten highly polarized word associations in both directions with computed z-score in Figure~\ref{fig:years}. These are the strongest word associations by z-score with the exception that, since we are interested in changes in the focus of papers and patents and not in the well-known evolution of specific types/names of models being used, we skip the words of "machine learning model(s) neural network(s)".

\subsubsection*{\textbf{Data Availability}}
Instructions for downloading and creating datasets used is available at \url{https://anonymous.4open.science/r/surv-cv-DD3F/README.md}.

\subsubsection*{\textbf{Code Availability}}
Code for this project is available at \url{https://anonymous.4open.science/r/surv-cv-DD3F/README.md}.

\section{Inclusion and Ethics Statement}
The authors of this paper are a multi-racial, multi-gender team with a wide range of expertise, including AI, machine learning, computer vision, NLP, robotics, cognitive science, philosophy, community organizing, critical theory, and security. The range of identities and expertises strengthened this paper, allowing us to understand the many subfields and impacts of computer vision and develop rigorous annotation schemes. This paper contends directly with the ethics of computer vision, helping uncover the extend of surveillance applications of CV and connect these applications to research.

\section*{Acknowledgements}
We owe gratitude and accountability to the long history of work exposing the nature of surveillance and how technology shifts power, work primarily done by
communities at the margins. Myra Cheng is supported by an NSF Graduate Research Fellowship (Grant DGE-2146755) and Stanford Knight-Hennessy Scholars graduate fellowship. Pratyusha Kalluri is supported in part by an Open Phil AI Fellowship.

\bibliographystyle{ACM-Reference-Format}
\bibliography{latex/biblio}

\appendix

\renewcommand{\thetable}{A\arabic{table}}

\renewcommand{\thefigure}{A\arabic{figure}}

\setcounter{figure}{0}

\setcounter{table}{0}

\section*{Appendix}

\section{The targeting of human data in papers versus patents}
\label{sec:data-extra}

\begin{figure*}[hbt!]
\caption{\textbf{The extraction of human data in computer vision papers versus downstream patents.} The breakdown for papers (\textit{n=100}) is as follows: human body parts (percent=35\% SD=4.7\%), human bodies (percent=36\% SD=4.7\%), human spaces (percent=18\% SD=3.8\%), salient traces (percent=1\% SD=0.8\%), unspecified (percent=9\% SD=2.8\%), and non-human data (percent=1\% SD=0.8\%). The breakdown for patents (\textit{n=100}) is as follows: human body parts (percent=27\% SD=4.5\%), human bodies (percent=38\% SD=4.9\%), human spaces (percent=16\% SD=3.8\%), salient traces (percent=5\% SD=2.2\%), unspecified (percent=13\% SD=3.3\%), and non-human data (percent=1\% SD=0.8\%). }
\vspace{10pt}
\centering
\includegraphics[width=.9\textwidth, trim={-1cm 3.2cm 1cm 0}, page=8]{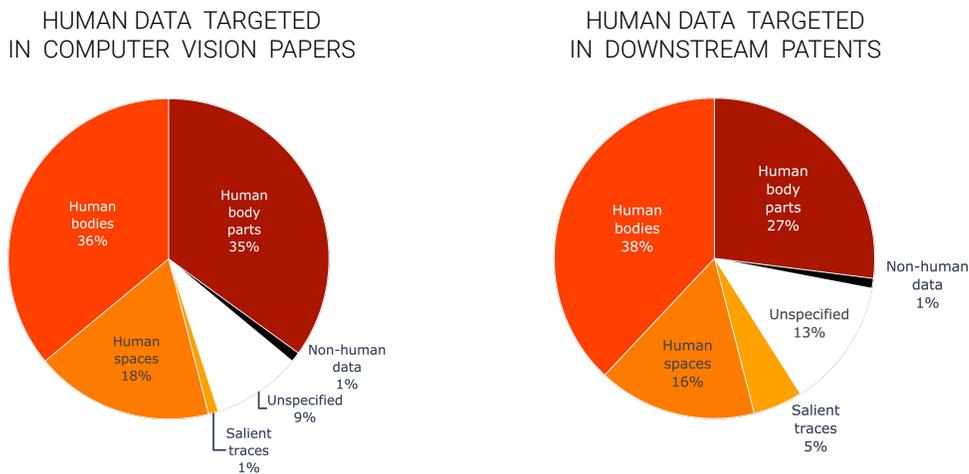}
\label{fig:data-type-split}
\end{figure*}

In Figure~\ref{fig:data-type-split} we present a quantitative summary of the types of data targeted in computer vision papers compared to downstream patents. We find similar trends. On the basis of our in-depth content analysis, we find that 90\% of papers and 86\% of patents extracted data relating to humans. Furthermore, the majority (71\% of papers and 65\% of patents) explicitly extracted data about human bodies and body parts. In some cases, papers that do not speak to the potential for use with human data led to patents that explicitly report monitoring human data. In general it was more common that papers did speak to targeting human data, and in particular targeting human body parts, and then patents leveraged these papers for overall targeting of human data or targeting of now unnamed data types. Only 1\% of papers and 1\% of patents were dedicated to targeting non-human data, showing that both computer vision research and applications are similarly concerned with analyzing, tracking, and monitoring humans and specifically human bodies.

\section{The transfer of human data}
\label{sec:transfer}
An additional central, organizing feature of surveillance is the mass collection, permanent storage, aggregation, and sharing of data, frequently without consent or awareness by the target individual, group or community~\cite{browne2015dark}. Regulatory bodies such as Europe's General Data Protection Regulation (GDPR)~\cite{EUdataregulations2018} and the California Consumer's Privacy Act of 2018 (CCPA)~\cite{de2018guide} have aimed to put mechanisms and regulations in place to ensure and enforce individual and collective privacy rights. GDPR outlines \textit{fair}, \textit{lawful} and \textit{transparent} data collection practices~\cite{malgieri2020concept}, deeming much of the current ubiquitous and aggressive nonconsensual mass data collection, transferal and sharing by surveillance companies/technologies unlawful. Subsequently, surveillance companies such as Clearview AI~\cite{ICO2022} as well as TikTok and Meta~\cite{ICOTikTok2022} are often found in breach of these data protection rights and face fines. % 
European data  regulation  authorities % 
for example, issued nearly €3bn in fines in 2022 alone~\cite{IrishTimes23}. Still,  % 
problematic and unlawful data collection, sharing, and transferal practices % 
have become the norm. From targeted online ads to wide ranging services (including, insurance, retail and finance) to ``smart'' home devices, future prediction is a core objective of surveillance technology~\cite{zuboff2019age}, which heavily relies on the vigorous collection, aggregation and transferal of data. Many studies of public attitudes reveal intense concern alongside a need for knowledge regarding the practices of data transferal.

We identified four categories capturing technologies' transferal of human data: 
\textit{the paper or patent anticipates transferring the data on a wireless connection; the data is transferred to another person or institution; the data is kept entirely locally; and whether and where data is stored or transferred is left ambiguous.} We found that stating data transferal, storage or management information is rarely mentioned in papers but relatively more common and conveyed in patents.

\begin{figure*}[!htb]
\caption{\textbf{The topology of Surveillance AI.} We present a topology of key dimensions of Surveillance AI. In particular, the relationship between complex technologies and surveillance can be clarified by attending to the \textit{extraction of human data}, the \textit{practices of data transferal}, and the \textit{institutional uses}. At each stage, we identify and describe prominent variants. In Section~\ref{sec:data} we present textual examples and analysis and quantify the  prevalence of these features of Surveillance AI.}
\centering

\includegraphics[width=1\textwidth, trim={0 2cm 0 0}, page=2]{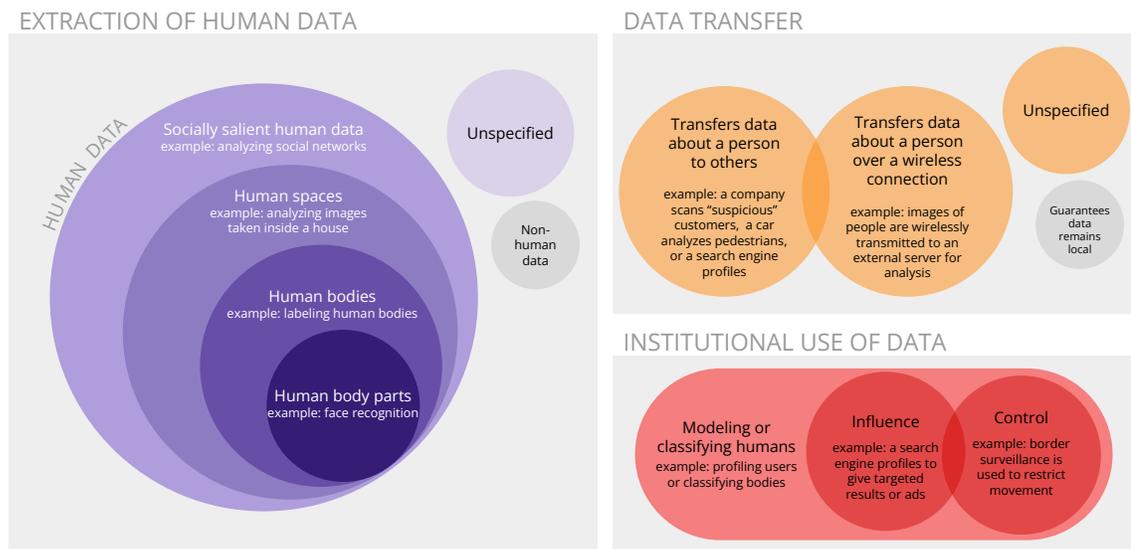}
\label{fig:topology}
\end{figure*}

\begin{center}\textbf{Data is transferred to others}    

\textit{``We developed methods for face recognition from sets of images...of the same unknown individual'' (Patent 0)}
\end{center}
\vspace{-.2em}
This category captured scenarios in which data about a person is not guaranteed to remain solely with that person and may instead be transferred to one or more other persons or institutions.
An example of this is a home video surveillance system that gives the system administrator access to videos of other persons, and may also share those videos with the manufacturer or other entities, such as law enforcement. In a world of `data economy'~\cite{o2017weapons} where AI systems are hungry for data,  data collected from our digital devices, fitness tracking technology and cameras provide insights about ourselves as well as our surroundings~\cite{gilliard2020caught}. Rarely, if at all, such data remains under the control of the data subject and is shared with third parties; institutes, data brokers, or other persons. Even when privacy polices are outlined, data is not guaranteed to remain under the control of the person. Examining 211 diabetes apps,~\citet{blenner2016privacy}, for example, found that of apps with privacy policies, 79 percent shared data while only about half of them admitted doing so. 
Similarly, a recent review of the privacy and data sharing policies of IoT devices and apps, found that despite restrictions in privacy policies, personal data is aggressively collected, shared and sold to third parties~\cite{mozilla22}.  

\begin{center}
\textbf{Data transfer over a wireless connection}      

\textit{``image data...may not be saved in intermediate form, but may simply be “piped” \\to a next stage over a bus, cable, wireless signal or other information channel'' (Patent 5)}
\end{center}
\vspace{-.2em}
\noindent Some patents indicate that image or video analysis will be done in the cloud and illustrate this in diagrams outlining their system.
Others do not explicitly mention that their artifact will be used to transfer data to an institution, but described the wireless capabilities of their artifact.
In both of the described scenarios we understand these as having the fully and intentionally anticipated capability for wireless data transfer. The collection, aggregation and categorization of data is one of the key characteristics of surveillance and an increasingly lucrative business~\cite{zuboff2019age,browne2015dark}. Even while appearing everyday and seemingly benign to many, ubiquitously connected technologies are instrumental for documenting, mapping, monitoring and facilitating widespread, networked surveillance. The under-regulated data broker industry and analytics companies, who infer individual features from consumer data in order to predict behaviour are an essential component of the surveillance ecosystem~\cite{reviglio2020datafied,west2019data,veliz2021privacy}. And, despite diverse understandings of the ideal that ought to be possible with internet and connectivity, in reality all connectivity serves an, at times shockingly productive, venue for data collection, aggregation, analytics, prediction, and ultimately surveillance~\cite{zuboff2019age}. According to Zuboff, ``Every avenue of connectivity serves to bolster private power’s need to seize behavior for profit.''

\begin{center}\textbf{Data remains exclusively local}\end{center}% 
\noindent Surveillance is not mere designing, building and deploying technologies, but is also marked by the struggle for power and control. A tracking technology such as a health monitor, for example, that exclusively remains under the control of a particular person, might serve only that particular user. This can potentially include papers and patents where all data collected is guaranteed to be kept and processed entirely at the control of the data subject, for example, on a personal server. Because this is entirely possible, we included this deductive code: the inclusion of this category served to actively search for and document any possible technology aimed placing total agency in the hands of the end user; however, \textit{none of the papers or patents fell into this category}.

\begin{center}
\textbf{Unspecified}
\end{center}
\vspace{-.5em}
Data transferal or storage information is sometimes undisclosed in patents and is rarely stated in papers. Note that this label does not prevent or limit any data from being transferred to others. Instead, it means that the work does not specify where or how the data is stored, shared, or transferred. Given that surveillance technologies tend to operate in the dark where technology vendors take extra measurements to hide their existence~\cite{hill2020secretive,brewster}, opacity in these category of papers and patents can signify purposeful obfuscation.

\section{The institutional use of data}\label{data-use-text}
\label{sec:app-use}
Surveillance is not mere passive observation but also extends on some capacity to control, regulate, or modulate behaviour~\cite{monahan2018surveillance}. These can be seemingly invisible influences, for example limiting choices or opportunities or directing people towards certain decisions (and away from others) through, for example, recommendation or personalization tools developed by big corporations. This form of influence and behaviour modulation is subtle and at times not recognized. Other times, papers and patents present a relatively direct surveillance application of their technology, where data is transferred and controlled by institutions, such as state and military bodies for the purpose of exercising power. Typically, surveillance technologies and norms are implemented and practised as convenience and a solution to “efficiency, productivity, participation, welfare, health or safety” whereby social control is framed as an unintended consequence~\cite{browne2015dark}. We identify three key data uses described in papers and patents.

\begin{center}\textit{\textbf{Modeling or categorizing humans}}\end{center}
The methods proposed in these works attempt to make humans amenable to modeling and categorization.
This form of surveillance might be to used collected data to generate models of humans without specifying the intended use case for these models.
An example of this could be pedestrian detection with no explicit purpose. Alternatively this form of surveillance might explicitly model and categorize to facilitate soft influence or hard control of humans.

\begin{center}

\textbf{Soft influence }\\\textit{``Applications including...real-time language translation, \\online search optimizations, and personalized user recommendations'' (Patent 35)}
\end{center}
Soft influence includes online targeted ads, recommendations, and other forms of personalization.
While surveillance used to exert soft influence does not require people to carry out certain behaviors, it can \textit{coerce} them towards behaviors that the surveillor believes are desirable and constraining the other options available to them.
An example of this is gaze tracking in mixed reality environments for targeted advertising. 
Note that while we use the term ``soft'' relative to the category of ``hard control,'' such influences can have significant, life-changing harms. For instance, personalized recommendations and advertisements can and have targeted vulnerable populations with misleading products and excluded marginalized communities from opportunities for employment, credit, and housing \citep{o2017weapons}.

\begin{center}
\textbf{Hard control}

\textit{``Applications include...assisting in automated patrol of large uncontrolled border crossing areas, \\such as the border between Canada and the US and/or the border between Mexico and the US.'' (Patent 5)}
\end{center}
Patents (more so than papers) often state exemplar applications of their technologies where data is transferred to institutions of power.
Examples of applications mentioned include tracking and identifying people for the purpose of border surveillance (for example, restricting movement), recognizing and tracking vehicles in cluttered urban scene using autonomous drones for the purpose of law enforcement, and detecting ``anomalous'' and suspicious activities.

\section{Additional details on the normalization of Surveillance AI}\label{sec:addtlinstitutions}

In Figure~\ref{fig:percents-labeled}, we present plots identical to those in Figure~\ref{fig:institutions} bottom with the addition of individual institutions', countries', and subfields' names listed. This enables closer inspection of, e.g. how specific institutions, nations, or subfields of interest participate in the contribution to surveillance. 

{

\begin{figure*}
\caption{\textbf{Additional details on the fieldwide dominance of downstream surveillance.} \\
This figure is identical to Figure~\ref{fig:institutions} bottom with the addition of individual institutions, countries, and subfields listed, enabling exploration of, e.g., which particular entities commit most intensively to surveillance. Best viewed as a PDF and zoomed in on. We identify each institution, nation, and subfield that has published at least 10  papers with downstream patents, and we compute the percent of these papers that are used in surveillance patents \textit{(vertical grey bars)} (\textit{n=13,804 papers}, \textit{n=18,272 papers}, \textit{n=19,413 papers}). In the case of subfields, rather than showing the large number of such subfields, we sort by the number of published computer vision papers and show only the first 150 subfields (\textit{n=18,946 papers}). As is described in the main text, we see a pervasive norm: \textit{when an institution, nation, or subfield authors papers with downstream patents, the majority are used in surveillance patents}. (\textit{Vertical grey bars are consistently above the orange 50\% threshold}.) Whiskers represent standard deviation.}
\centering
\vspace{.5cm}
\includegraphics[width=150px, trim={0 7cm  0cm 0}, clip, center]{latex/images/percent_legend_h.pdf}
\includegraphics[height=160px, trim={0cm 0 0 15cm}, clip]{latex/images/inst_percents_labeled.pdf}
\includegraphics[height=90px, trim={0cm 32cm 0 22cm}, clip]{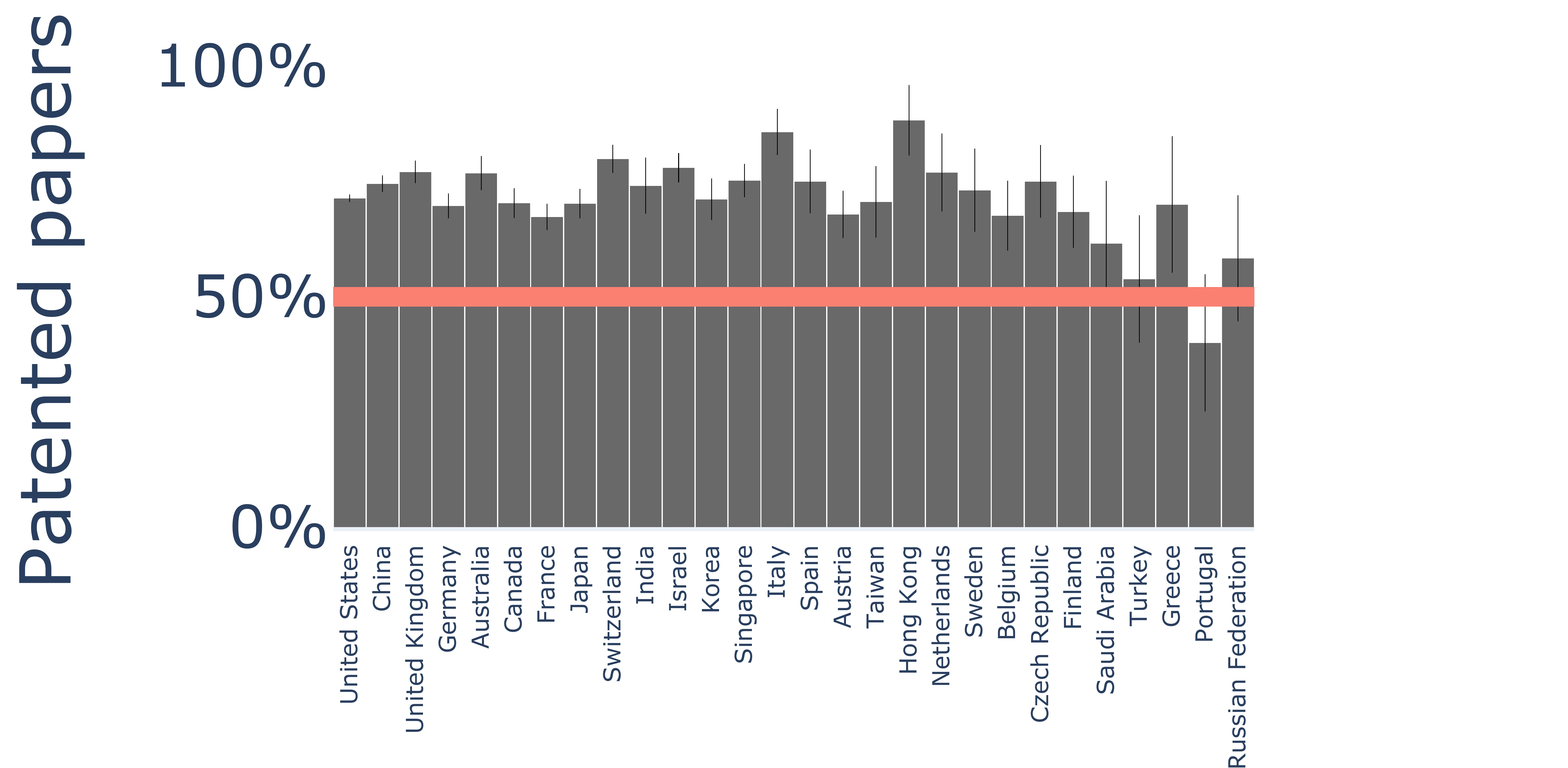} \hfill
\includegraphics[height=160px, trim={0cm 0 0 15cm}, clip]{latex/images/subfield_percents_labeled.pdf}
\label{fig:percents-labeled}
\end{figure*}
}

We especially draw attention to the case of subfields. Computer vision is a broad field comprised of several subfields, such as object detection, medical imaging, and 3D reconstruction. 
Although some research in these subfields is more recognizable as surveillance (e.g., facial recognition) than others, we found that research across many subfields has contributed to the creation of Surveillance AI. For 77\% of subfields that author papers with downstream patents, at least half of these papers are used in surveillance patents (\textit{vertical bars are above the orange 50\% threshold}).
As can be seen, there are several topics explicitly related to the tracking of human bodies, such as ``face detection'' and ``motion detection.'' Yet interestingly, many of the top subfields have no explicit relation to the modeling of human data and instead are simply common topics, like ``background subtraction'' and ``computer graphics.'' The wide distribution of topics across papers cited in surveillance patents reveal that the work of many subfields, even ones not explicitly connected to human data, have contributed to Surveillance AI.

\section{Additional background on Surveillance and Computer Vision} \label{sec:background}

\subsubsection*{\textit{\textbf{Surveillance}}}

Surveillance is a technology of social control intrinsically tied to the production of power relations. The observing, tracing and monitoring is practised by those in a relative position of power to those being observed. The enactments of surveillance frequently reify boundaries, borders, and bodies along racial lines, the consequence of which is often discriminatory treatment of individuals and communities that are negatively portrayed, which Browne terms ``Racializing surveillance''~\cite{browne2015dark}. Surveillance perpetually influences its subjects in making them more ``amenable to observations, prediction, and suggestion''~\cite{cohen2017surveillance}. Surveillance practices produce social norms and standards and exercise the ``power to define what is in or out of place''~\cite{browne2015dark}.    % 
State powers and military institutions track, monitor and profile citizens, immigrants, ``offenders'' or ``suspects''; companies watch and monitor their employees; tech corporations track, sort and profile users; education institutes track and monitor their pupils, often with the justification of enhancing security, productivity, safety, or efficiency. As most surveillance technologies are designed, developed, and deployed by and for institutions of power as the paying customers and primary stakeholders, the safety, welfare, and interest of individuals and communities where these technologies are deployed are an afterthought, if considered at all. As a result, while institutions of power benefit the most from the production and deployment of surveillance technologies, communities subjected to surveillance (often communities at the margins of society) are disproportionately negatively impacted~\cite{benjamin2019race,browne2015dark,lyon2010surveillance,eubanks2018automating}. Current facial recognition technologies used in law enforcement, for example, disproportionately negatively impact racial minorities. In the context of US law enforcement, for example, facial recognition surveillance has so far led to at least four wrongful arrest,  all of whom are Black men~\cite{germain23}, facilitating and expanding racialized carceral systems. Relational and collective conceptions of surveillance are therefore critical for a comprehensive understanding of surveillance that can account for power asymmetries that permeate the surveillance ecology.

\subsubsection*{\textit{\textbf{Computer Vision}}} 

The emergence of the World Wide Web and with it, the `availability' of vast amount of image and video data, has been a central contributing factor for the rapid rise of the field in the past decade. Critiques have been formulated that draw attention not only to the histories, but the encoded values and ongoing practices within the field. % 
Dataset collection, curation and management practices, in most cases remain devoid of careful considerations of issues such as informed consent, privacy, or dataset audits (for example, to mitigate negative social stereotypes often encoded in data)~\cite{peng2021mitigating,birhane2021large}. Dataset collection and curation practices in computer vision are compared to the ethical equivalent of data theft~\cite{paullada2021data} and erode privacy with most data collected without informed consent or procedures to opt-out~\cite{paullada2021data,birhane2021large}. Dataset collection, documentation, and development in computer vision are driven by the underlying values of efficiency, universality, impartiality, and model work. \citet{scheuerman2021datasets} et al. further note that ``Efficiency is valued over care, a slow and more thoughtful approach to dataset curation. Universality is valued over contextuality, a focus on more specific tasks, locations, or audiences. Impartiality is valued over positionality'' based on extensive analysis of canonical image datasets. Furthermore, the rapid rise and accessibility of generative models such as Stable Diffusion, not only exacerbates the erosion of privacy and the reproduction of social stereotypes, toxic and discriminatory predictions, the proliferation  of these generated images at a massive scale also pollutes the digital ecology~\cite{bianchi2022easily}. 
Against the backdrop of the rapid rise of computer vision and the growing array of critiques of the field, it is crucial we understand the extent and nature of the flow from the field's histories, values, and practices to major downstream applications such as surveillance.

\section{Additional methodological details}
\label{sec:method-extra} 

\subsection{Automated analysis}
\label{ref:keywords}
We searched the abstract and body of each patent for the following surveillance indicator words: ``ad'', ``advertisement'', ``airport'', ``apartment'', ``army'', ``baggage'', ``caste'', ``citizen'', ``combat'', ``convict'', ``crime'', ``criminal'', ``defense'', ``disability'', ``enemy'', ``ethnicity'', ``face'', ``facial'', ``facial recognition'', ``felon'', ``female'', ``foot traffic'', ``fraud'', ``friend'', ``gender'', ``geolocation'', ``hand'', ``iris'', ``irises'', ``jail'', ``kid'', ``license plate'', ``limb'', ``male'', ``man'', ``military'', ``nonbinary'', ``office'', ``pedestrian'', ``penitentiary'', ``prison'', ``prisoner'', ``purchase'', ``recommend'', ``reidentification'', ``security'', ``sex'', ``sexuality'', ``social network'', ``street'', ``surveil'', ``surveillance'', ``torso'', ``transgender'', ``underage'', ``woman'', and ``youth''.

While we in general searched the patent abstracts and patent bodies, limitations of our parser resulted in, for a small number of patents, being able to obtain and search only the abstract. Our keyword counts thus constitute lower bounds, and the prevalence of surveillance is likely to be even greater than that which we document.

\subsection{Error Estimation}

For any quantitative result from our manual and automated analysis, we report standard deviation. 
Standard deviation is estimated using a boostrapping algorithm. 
For any given subset of $n$ papers or patents, we sample $k=1,000$ sets of the $n$ elements with replacement, and calculate the standard deviation of positive and negative classes.
For the manual coding, an item is in the positive class if it has been annotated as such (\textit{i.e.}, if it falls under one of the 4 categories shown in Table~\ref{tab:examples}).
For the automated analysis, an item is in the positive class if it is associated with any of the keywords listed in Appendix~\ref{ref:keywords}.

\end{document}